# Cross-Modal Retrieval with Implicit Concept Association

Yale Song
Microsoft AI & Research
yalesong@microsoft.com

Mohammad Soleymani
University of Geneva
mohammad.soleymani@unige.ch

## ABSTRACT

Traditional cross-modal retrieval assumes explicit association of concepts across modalities, where there is no ambiguity in how the concepts are linked to each other, e.g., when we do the image search with a query "dogs", we expect to see dog images. In this paper, we consider a different setting for cross-modal retrieval where data from different modalities are *implicitly* linked via concepts that must be inferred by high-level reasoning; we call this setting implicit concept association. To foster future research in this setting, we present a new dataset containing 47K pairs of animated GIFs and sentences crawled from the web, in which the GIFs depict physical or emotional reactions to the scenarios described in the text (called "reaction GIFs"). We report on a user study showing that, despite the presence of implicit concept association, humans are able to identify video-sentence pairs with matching concepts, suggesting the feasibility of our task. Furthermore, we propose a novel visual-semantic embedding network based on multiple instance learning. Unlike traditional approaches, we compute multiple embeddings from each modality, each representing different concepts, and measure their similarity by considering all possible combinations of visual-semantic embeddings in the framework of multiple instance learning. We evaluate our approach on two video-sentence datasets with explicit and implicit concept association and report competitive results compared to existing approaches on cross-modal retrieval.

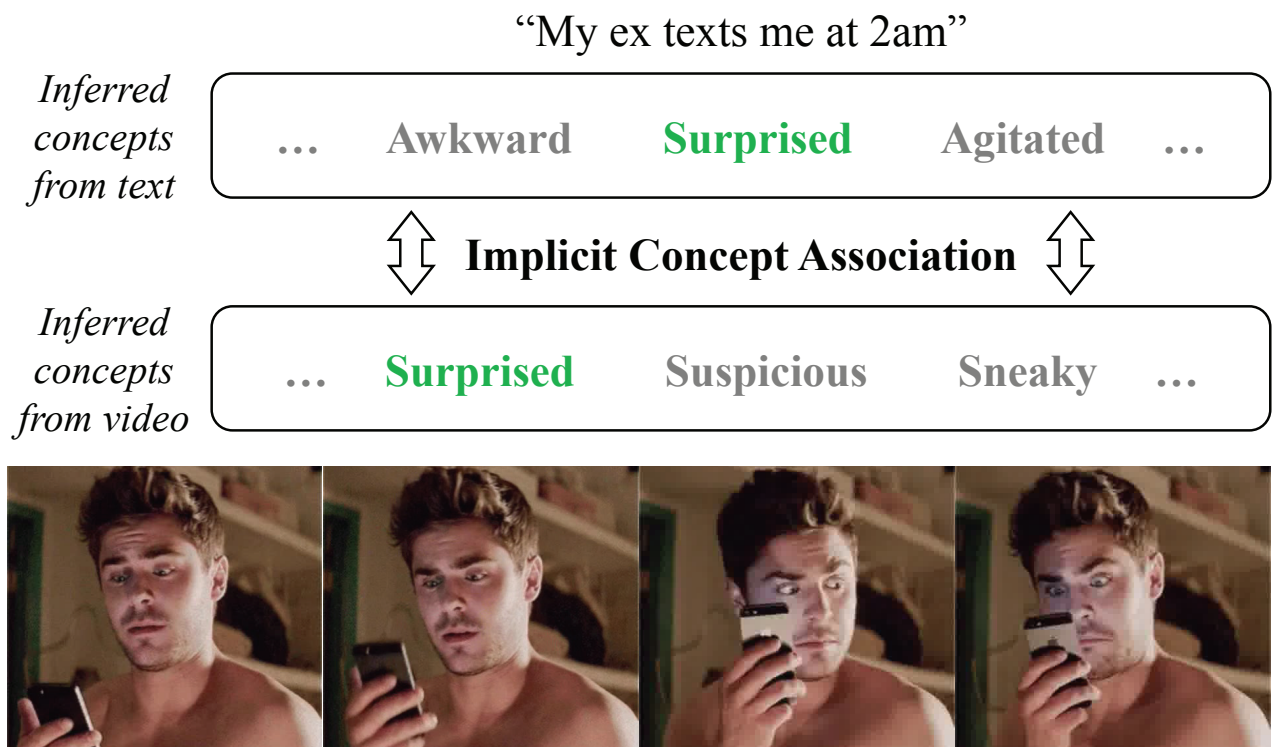

**Figure 1: We consider a scenario for cross-modal retrieval where data from different modalities are implicitly linked via "concepts" (e.g., emotion, intent) that must be inferred by high-level reasoning. Each modality may manifest multiple different concepts; only a small subset of all visual-textual concept pairs could be related (green), while others are irrelevant (gray). This scenario poses new challenges to cross-modal retrieval.**

## 1 INTRODUCTION

Animated GIFs are becoming increasingly popular [3]; more people use them to tell stories, summarize events, express emotion, and enhance (or even replace) text-based communication. To reflect this trend, several social networks and messaging apps have recently incorporated GIF-related features into their systems, e.g., Facebook users can create posts and leave comments using GIFs, Instagram and Snapchat users can put "GIF stickers" into their personal videos, and Slack users can send messages using GIFs. This rapid increase in popularity and real-world demand necessitates more advanced and specialized systems for animated GIF search. The key to the success of GIF search lies in the ability to understand the complex semantic relationship behind textual queries and animated GIFs, which, at times, can be implicit and even ambiguous. Consider a query term *"my ex texts me at 2am"*: How can we build a system that understands the intent behind such queries and find relevant content, e.g., GIFs showing someone surprised or agitated?

In this paper, we focus on cross-modal retrieval when there is *implicit* association of concepts across modalities. Our work is in part motivated by the above scenario (see also Figure 1), where the semantic relationship between textual queries and visual content is implied only indirectly, without being directly expressed. In the example above, the two modalities are related with such concepts as "surprised" and "panicked" – these are abstract concepts that are not explicitly stated or depicted in either modalities; they are rather something that must be inferred from data by high-level reasoning. In this regard, we refer to our task *cross-modal retrieval with implicit concept association*, where "implicit concepts" are implied and not plainly expressed in the data.

Previous work on cross-modal retrieval has mainly focused on the scenario of *explicit* concept association, in which there is no ambiguity in how concepts are linked to each other [28, 37]. For example, most popular and widely accepted visual-textual retrieval datasets, such as NUS-WIDE [7], Flickr30K [42], and MS-COCO [24], contain user tags and sentences describing visual objects and scenes displayed explicitly in the corresponding images. Also, the concept vocabulary used in those datasets have clear image-to-text correspondence (a dog image is tagged with the "dog" category), leaving less room for ambiguity. Furthermore, most existing approaches to cross-modal retrieval focus on learning a shared embedding space where the distance between data from two modalities are minimized, e.g., DCCA [2], DeViSE [13], and Corr-AE [12]; all these methods are however based on an assumption that there is explicit association of concepts across modalities. To the best of our knowledge, there is no literature on the case with implicit concept association.

MRW a witty comment I wanted to make was already said | MFW I see a cute girl on Facebook change her status to single | MFW I can't remember if I've locked my front door | MRW a family member asks me why his computer isn't working

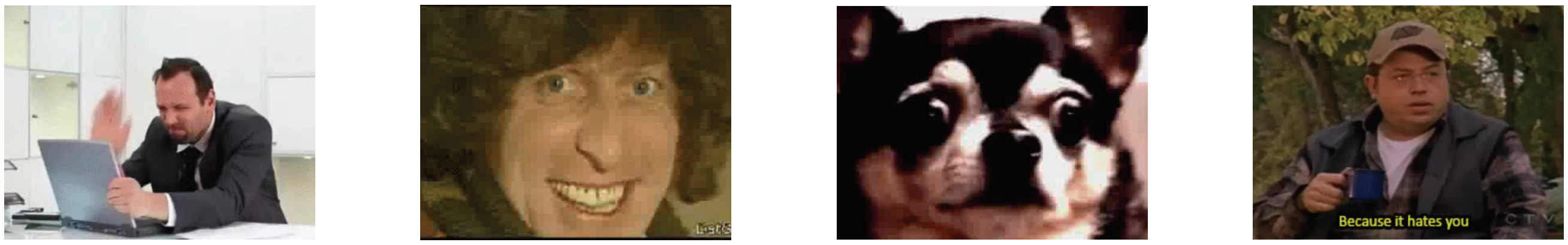

(a) Physical Reaction | (b) Emotional Reaction | (c) Animal Reaction | (d) Lexical Reaction (Caption)

**Figure 2: Our MRW dataset contains 47K pairs of GIF-sentence pairs, where GIFs depict reactions to the sentences (often called "reaction GIFs"). Here we show the four most common types of reactions: (a) physical, (b) emotional, (c) animal, (d) lexical.**

In this work, we make a step toward cross-modal retrieval with implicit concept association. To foster further research in this direction, we collect a new dataset containing 47K pairs of animated GIFs and sentences, where the GIFs depict physical or emotional reactions to certain situations described in text – they are often called "reaction GIFs". We name our dataset MRW (my reaction when) following the most popular hashtag for reaction GIFs. To highlight the difference between implicit vs. explicit concept association, we compare our dataset with the TGIF dataset [23] that also contains GIF-sentence pairs, originally proposed for video captioning. In the TGIF dataset, sentences describe objects and actions explicitly displayed in visual content; it can, therefore, be considered as explicit concept association. We report on a user study that shows, despite the presence of implicit concept association in our dataset, humans are still able to identify GIF-sentence pairs with matching concepts with a high confidence, suggesting the feasibility of our task.

Furthermore, we propose a novel approach to visual-semantic embedding based on multiple instance learning; we call our model **M**ultiple-**i**nstance **Vi**sual-**S**emantic **E**mbedding (MiViSE) networks. To deal with implicit concept association, we formulate our learning problem as many-to-many mapping, in which we compute several different representations of data from each modality and learn their association in the framework of multiple instance learning. By considering all possible combinations of different representations between two modalities (i.e., many-to-many), we provide more flexibility to our model even in the presence of implicit concept association. We employ recurrent neural networks [6] with self attention [25] to compute multiple representations, each attending to different parts of a sequence, and optimize the model with triplet ranking so that the distance between true visual-textual pairs are minimized compared to negative pairs. We evaluate our approach on the sentence-to-video retrieval task using the TGIF and our MRW datasets, and report competitive performance over several baselines used frequently in the cross-modal retrieval literature.

In summary, we make the following contributions:

(1) We focus on cross-modal retrieval with implicit concept association, which have not been studied in the literature.
(2) We release a new dataset MRW (my reaction when) that contains 47K pairs of animated GIFs and sentences.
(3) We propose a novel approach to cross-modal retrieval based on multiple instance learning, called MiViSE.

## 2 RELATED WORK

**Datasets**: Existing datasets in cross-modal retrieval have different types of visual-textual data pairs, categorized into: *image-tag* (e.g., NUS-WIDE [7], Pascal VOC [11], COCO [24]), *image-sentence* (e.g., Flickr 8K [30], Flickr 30K [42], and COCO [24]), and *image-document* (e.g., Wikipedia [31] and Websearch [22]).

Unlike existing datasets, ours contains video-sentence pairs; it is therefore more closely related to video captioning datasets, e.g., MSR-VTT [39], TGIF [23], LSMDC [32]. All those datasets contain visual-textual pairs with explicit concept association (sentences describe visual content displayed explicitly in videos), whereas ours contains visual-textual pairs with implicit concept association (videos contain physical or emotional reactions to sentences).

**Visual-semantic embedding**: The fundamental task in cross-modal retrieval is finding an embedding space shared among different modalities such that samples with similar semantic meanings are close to each other. One popular approach is based on maximizing correlation between modalities [2, 9, 12, 14, 31, 41]. Rasiwasia *et al*. [31] use canonical correlation analysis (CCA) to maximize correlation between images and text, while Gong *et al*. [14] propose an extension of CCA to a three-view scenario, e.g., images, tags, and their semantics. Following the tremendous success of deep learning, several work incorporate deep neural networks into their approaches. Andrew *et al*. [2] propose deep CCA (DCCA), and Yan *et al*. [41] apply it to image-to-sentence and sentence-to-image retrieval. Based on an idea of autoencoder, Feng *et al*. [12] propose a correlation autoencoder (Corr-AE), and Eisenschtat and Wolf [9] propose 2-way networks. All these methods share one commonality: They aim to find an embedding space where the correlation between pairs of cross-modal data is maximized.

Besides the pairwise correlation approaches, another popular approach is based on triplet ranking [13, 38]. The basic idea is to impose a constraint that encourages the distance between "positive pairs" (e.g., ground-truth pairs) to be closer than "negative pairs" (typically sampled by randomly shuffling the positive pairs). Frome *et al*. [13] propose a deep visual-semantic embedding (DeViSE) model, using a hinge loss to implement triplet ranking. Similar to DeViSE, we also train our model using triplet ranking. However, instead of using the hinge loss of [13], which is non-differentiable, we use the pseudo-Huber loss form [4], which is fully differentiable and therefore more suitable for deep learning. Through ablation

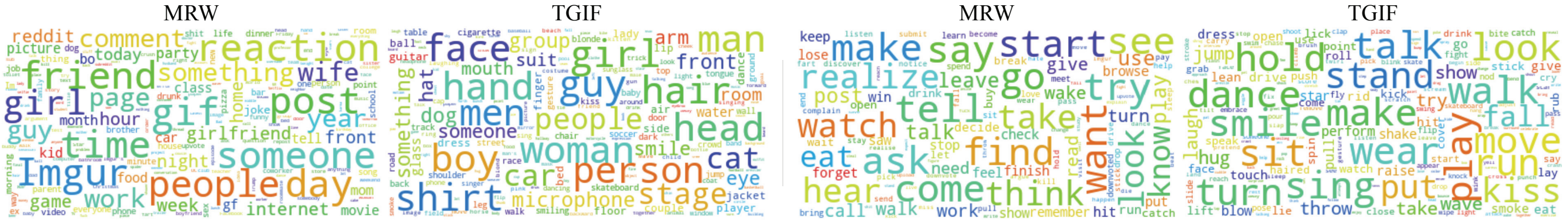

(a) Nouns

(b) Verbs

**Figure 3: Distributions of nouns and verbs in our MRW and the TGIF [23] datasets. Compared to the TGIF, words in our dataset depict more abstract concepts (e.g., post, time, day, start, realize, think, try), suggesting the implicit nature in our dataset.**

studies we show improved performance just by replacing the hinge loss to the pseudo-Huber loss.

Recent approaches to cross-modal retrieval apply techniques from machine learning, such as metric learning and adversarial training [35, 36]. Tasi *et al*. [35] train deep autoencoders with maximum mean discrepancy (MMD) loss, while Wang *et al*. [36] combine the triplet ranking-based method with adversarial training. Unlike all previous approaches, our work focuses specifically on the scenario of implicit concept association, and train our model in the framework of multiple instance learning (MIL). To the best of our knowledge, our work is the first to use MIL in the cross-modal retrieval setting.

**Animated GIF**: There is increasing interest in conducting research around animated GIFs. Bakhshi *et al*. [3] studied what makes animated GIFs engaging on social networks and identified a number of factors that contribute to it: the animation, lack of sound, immediacy of consumption, low bandwidth and minimal time demands, the storytelling capabilities and utility for expressing emotions.

Several work use animated GIFs for various tasks in video understanding. Jou *et al*. [19] propose a method to predict viewer perceived emotions for animated GIFs. Gygli *et al*. [16] propose the Video2GIF dataset for video highlighting, and further extended it to emotion recognition [15]. Chen *et al*. [5] propose the GIFGIF+ dataset for emotion recognition. Zhou *et al*. [44] propose the Image2GIF dataset for video prediction, along with a method to generate cinemagraphs from a single image by predicting future frames.

Recent work use animated GIFs to tackle the vision & language problems. Li *et al*. [23] propose the TGIF dataset for video captioning; Jang *et al*. [18] propose the TGIF-QA dataset for video visual question answering. Similar to the TGIF dataset [23], our dataset includes video-sentence pairs. However, our sentences are created by real users from Internet communities rather than study participants, thus posing real-world challenges. More importantly, our dataset has *implicit* concept association between videos and sentences (videos contain physical or emotional reactions to sentences), while the TGIF dataset has *explicit* concept association (sentences describe visual content in videos). We discuss this in more detail in Section 3.

## 3 MRW DATASET

Our dataset consists of 47K pairs of GIFs and sentences collected from popular social media websites including reddit, Imgur, and Tumblr.[1] We crawl the data using the GIPHY API[2] with query terms `mrw`, `mfw`, `hifw`, `reaction`, and `reactiongif`; we crawled the data from August 2016 to January 2018. Figure 2 shows examples of GIF-sentence pairs in our dataset.

| | |
|---|---|
| Total number of video-sentence pairs | 47,172 |
| Median number of frames in a video | 48 |
| Median number of words in a sentence | 10 |
| Word vocabulary size | 3,706 |
| Average term frequency | 18.47 |
| Median term frequency | 1 |

**Table 1: Descriptive statistics of the MRW dataset.**

Our dataset is unique among existing video-sentence datasets. Most existing ones are designed for video captioning [23, 32, 39] and assume *explicit* association of concepts between videos and sentences: *sentences describe visual content in videos*. Unlike existing video-sentence datasets, ours focuses on a unique type of video-sentence pairs, i.e., reaction GIFs. According to a popular subreddit channel `r/reactiongif`[3]:

> *A reaction GIF is a physical or emotional response that is captured in an animated GIF which you can link in response to someone or something on the Internet. The reaction must not be in response to something that happens within the GIF, or it is considered a "scene".*

This definition clearly differentiates ours from existing video-sentence datasets: ours assume *implicit* association of concepts between videos and sentences, i.e., *videos contain reactions to sentences*.

We call our dataset MRW (my reaction when) to emphasize the unique characteristic of our data – MRW is the most popular hashtag for reaction GIFs.[4]

### 3.1 Data Analysis

Table 1 shows descriptive statistics of our dataset, and Figure 3 shows word clouds of nouns and verbs extracted from our MRW dataset and the TGIF dataset [23]. Sentences in the TGIF dataset are constructed by crowdworkers to describe the visual content explicitly displayed in GIFs. Therefore, its nouns and verbs mainly describe physical objects, people and actions that can be visualized, e.g., cat, shirt, stand, dance. In contrast, MRW sentences are constructed by the Internet users, typically from subcommunities in social networks that focus on reaction GIFs. As can be seen from Figure 3, verbs and nouns in our MRW dataset additionally include abstract terms that cannot necessarily be visualized, e.g., time, day, realize, think.

[1] https://www.reddit.com, https://imgur.com, https://www.tumblr.com
[2] https://developers.giphy.com
[3] https://www.reddit.com/r/reactiongifs
[4] 31,695 out of 47,172 samples (67%) in our dataset contain this hashtag.

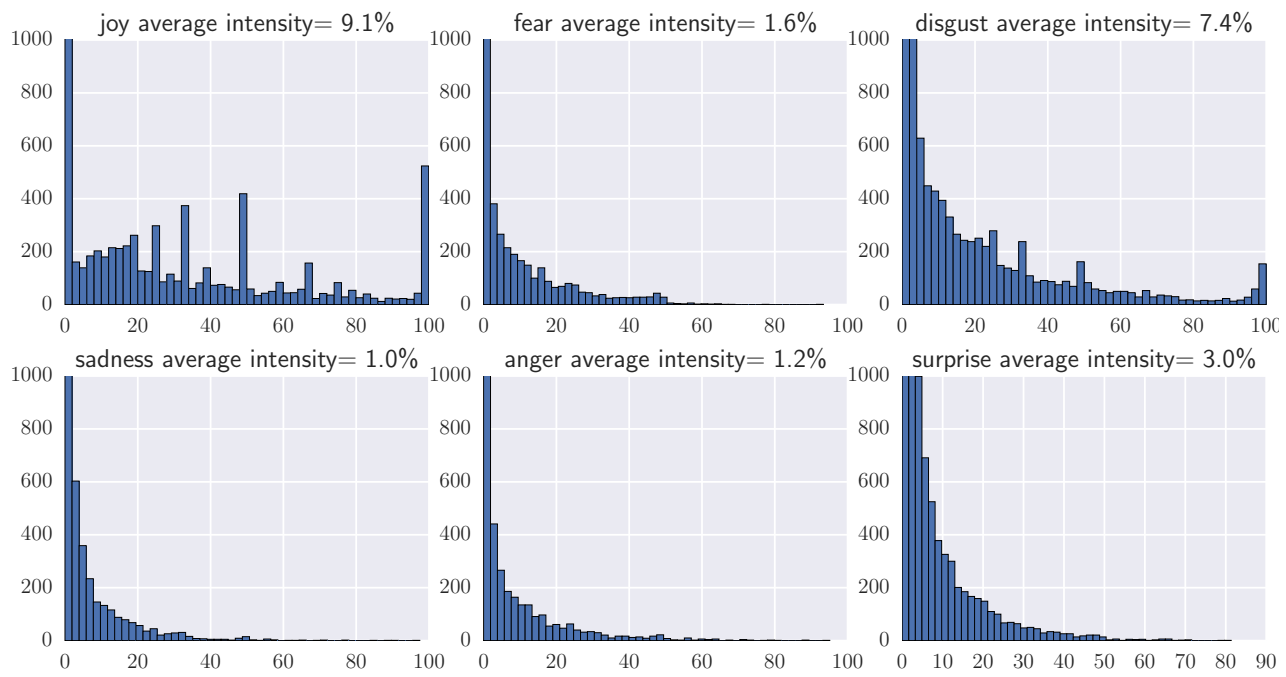

**Figure 4: Histograms of the intensity of facial expressions. The horizontal axis represents the intensity of the detected expression, while the vertical axis is the sample count in frames with faces. We clip the $y$-axis at 1000 for visualization. Overall, joy, with average intensity of 9.1% and disgust (7.4%) are the most common facial expressions in our dataset.**

Facial expression plays an important role in our dataset: 6,380 samples contain the hashtag MFW (my face when), indicating that those GIFs contain emotional reactions manifested by facial expressions. Therefore, we apply automatic facial expression recognition to analyze the types of facial expressions contained in our dataset.

First, we count the number of faces appearing in the GIFs. To do this, we applied the dlib CNN face detector [21] on five frames sampled from each GIF with an equal interval. The results show that there are, on average, 0.73 faces in a given frame of an animated GIF; 32,620 GIFs contain at least one face. Next, we use the Affectiva Affdex [26] to analyze facial expressions depicted in GIFs, detecting the intensity of expressions from two frames per second in each GIF. We looked at six expressions of basic emotions [10], namely, joy, fear, sadness, disgust, surprise and anger. We analyzed only the frames that contain a face with its bounding box region larger than 15% of the image. Figure 4 shows the results. Overall, joy with average intensity of 9.1% and disgust (7.4%) are the most common facial expressions in our dataset.

## 3.2 Explicit vs. Implicit Association: A User Study

Image and video captioning often involves describing objects and actions depicted explicitly in visual content [23, 24]. For reaction GIFs, however, visual-textual association is not always explicit. For example, objects and actions depicted in visual content might be a physical or emotional reaction to the scenario posed in the sentence.

To study how clear these associations are for humans, we conducted a user study in which we asked six participants to verify the association between sentences and GIFs. We randomly sampled 100 GIFs from the test sets of both our dataset and TGIF dataset [23]. We chose the test sets due to the higher quality of the data. We paired each GIF with both its associated sentence and a randomly selected sentence, resulting in 200 GIF-sentence pairs per dataset.

The results show that, in case of our dataset (MRW), 80.4% of the associated pairs are positively marked as being relevant, suggesting humans are able to distinguish the true vs. fake pairs despite implicit concept association. On the other hand, 50.7% of the randomly assigned sentences are also marked as matching sentences. The high false positive rate shows the ambiguous nature of GIF-sentence association in our dataset. In contrast, for TGIF dataset with clear explicit association, 95.2% of the positive pairs are correctly marked as relevant and only 2.6% of the irrelevant pairs are marked as being relevant. This human baseline demonstrates the challenging nature of GIF-sentence association in our dataset, due to their implicit rather than explicit association. This motivates our multiple-instance learning based approach, described next.

# 4 APPROACH

Our goal is to learn embeddings of videos and sentences whose concepts are implicitly associated. We formalize this in the framework of multiple instance learning with triplet ranking.

We assume a training dataset $\mathcal{D} = \{(V_n, S_n^+, S_n^-)\}_{n=1}^N$, where $V_n = [v_{n,1}, \cdots, v_{n,T_v}]$ is a video of length $T_v$ and $S_n^+ = [s_{n,1}^+, \cdots, s_{n,T_s}^+]$ is a sentence of length $T_s$; the pair $(V_n, S_n^+)$ has matching concepts. For each pair, we randomly sample a negative sentence $S_n^-$ from the training set for the triplet ranking setup. For notational simplicity, we drop the subscript $n$ and use $S$ to refer to both $S^+$ and $S^-$ unless a distinction between the two is necessary.

Our model (see Figure 5) includes a video encoder and a sentence encoder. Each encoder outputs $K$ representations, each computed by attending to different parts of a video or a sentence. We use the representations to train our model using a triplet ranking loss in the multiple instance learning framework. Below we describe each component of our model in detail.

## 4.1 Video and Sentence Encoders

Our video and sentence encoders operate in a similar fashion: once we compute image and word embeddings from video and sentence input, we use a bidirectional RNN with self-attention [25] to obtain $K$ embeddings of a video and a sentence, respectively.

We compute an image embedding of each video frame $v_t \in V$ using a convolutional neural network (CNN); we use the penultimate layer of ResNet-50 [17] pretrained on ImageNet [33], and denote its output by $x_t^v \in \mathbb{R}^{2048}$. Similarly, we compute a word embedding of each word $s_t \in S$ using the GloVe [29] model pretrained on the Twitter dataset, and denote its output by $x_t^s \in \mathbb{R}^{200}$.

Next, we process a sequence of image embeddings using the bi-directional Gated Recurrent Units (GRU) [6],

$$\overrightarrow{h_t^v} = \text{GRU}(x_t^v, \overrightarrow{h_{t-1}^v}), \quad \overleftarrow{h_t^v} = \text{GRU}(x_t^v, \overleftarrow{h_{t+1}^v}), \tag{1}$$

and concatenate the $\overrightarrow{\text{forward}}$ and $\overleftarrow{\text{backward}}$ hidden states to obtain $h_t^v = [\overrightarrow{h_t^v}, \overleftarrow{h_t^v}]$. We denote the resulting sequence of hidden states by

$$H^v = [h_1^v, \cdots, h_{T_v}^v] \tag{2}$$

which is a matrix of size $d$-by-$T_v$, where $d$ is the total number of GRU hidden units (forward and backward combined). We follow the same steps to process a sequence of word embeddings and obtain a matrix $H^s = [h_1^s, \cdots, h_{T_s}^s]$ of dimension $d$-by-$T_s$.

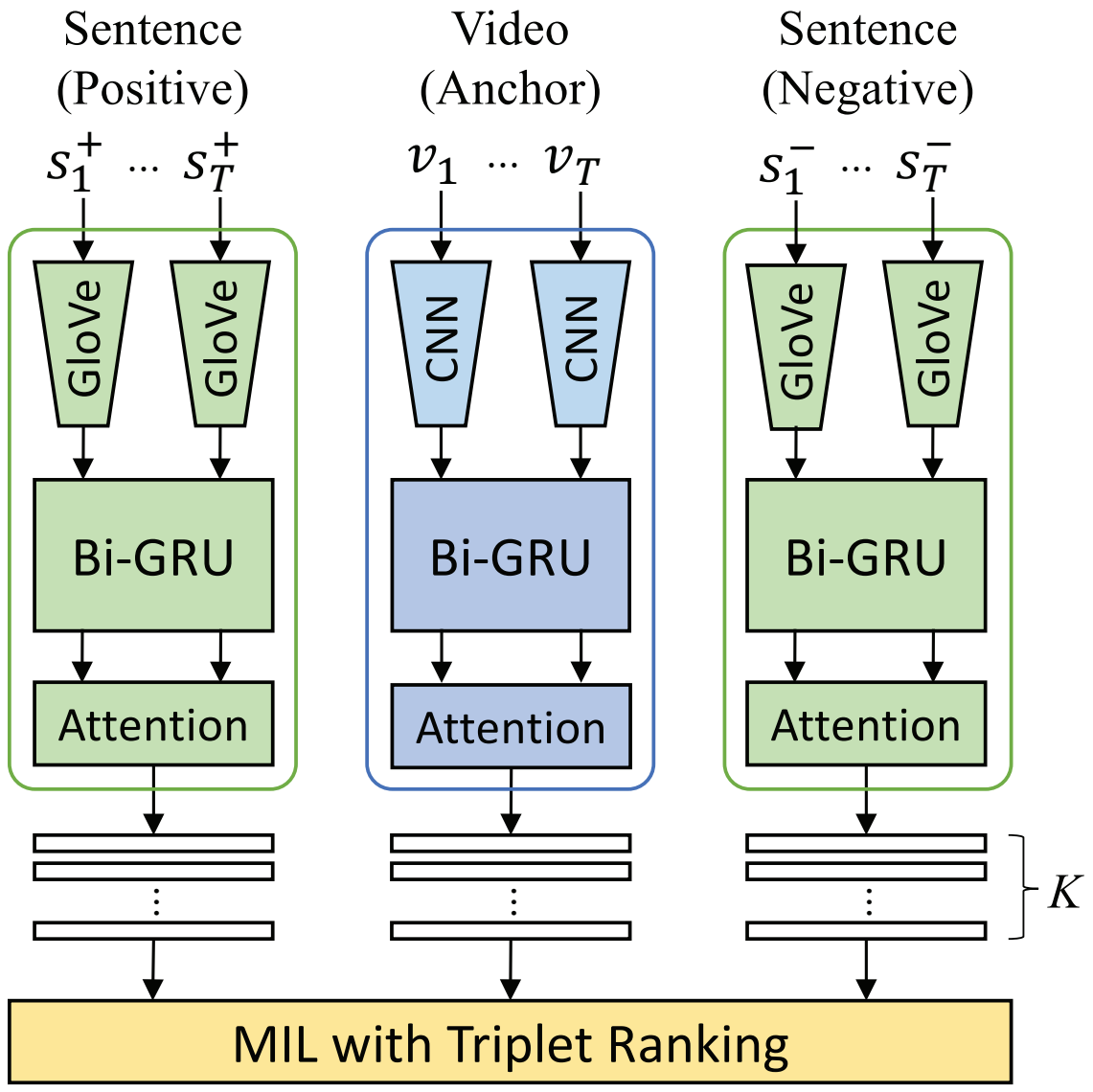

**Figure 5: Our model consists of a video encoder (blue) and a sentence encoder (green). Each encoder computes $k$ embeddings, each attending to different parts of a sequence of hidden states inferred from a bidirectional GRU. We train our model with our novel MIL-based triplet ranking formulation.**

Finally, we compute a linear combination of the hidden states to obtain $K$ video embeddings $\phi(V)_i$ and $K$ sentence embeddings $\psi(S)_j$, $i, j = 1...K$, where each embedding selectively attends to different parts of the sequence using self attention [25]. This is done by computing a self-attention map, e.g., for video input:

$$A^v = \texttt{softmax}\left(W_2^v \texttt{tanh}(W_1^v H^v)\right) \quad (3)$$

where $W_1^v$ is a weight matrix of size $u$-by-$d$ and $W_2^v$ is of dimension $K$-by-$u$; we set $u = d$ per empirical observation. The softmax is applied row-wise to obtain an attention map $A^v$ of dimension $K$-by-$T_v$, where each of $K$ rows sums up to 1. We then obtain $K$ video embeddings as a linear combination

$$\Phi(V) = A^v \left(H^v\right)^\intercal \quad (4)$$

where $\Phi(V) = [\phi(V)_1, \cdots, \phi(V)_k]$ is a matrix of size $K$-by-$d$ with each row serving as an embedding. We use the same method to obtain $K$ sentence embeddings $\Psi(S) = [\psi(S)_1, \cdots, \psi(S)_K]$.

Each of the $K$ embeddings selectively attends to different parts of a sequence, and thus represents different concepts contained in a video or a sentence. As we shall see later, having multiple embeddings allows us to formulate our problem in the MIL framework.

## 4.2 Triplet Ranking

We train our model with a triplet ranking loss, using $V$ as the anchor. The standard choice for triplet ranking is the hinge loss:

$$\mathcal{L}_{hinge}(V, S^+, S^-) = \max\left(0,\ \rho - \Delta(V, S^+, S^-)\right) \quad (5)$$

where $\rho$ sets the margin and $\Delta(\cdot)$ measures the difference between $(V, S^+)$ and $(V, S^-)$ with respect to a suitable metric; we define it as

$$\Delta(V, S^+, S^-) = f(V, S^+) - f(V, S^-) \quad (6)$$

where $f(V, S)$ measures the similarity between $V$ and $S$. This hinge loss form is non-differentiable at $\rho$, making optimization difficult to solve. Therefore, we adapt the pseudo-Huber loss formulation [4]:

$$\mathcal{L}_{huber}(V, S^+, S^-) = \delta^2 \left(\sqrt{1 + ((\rho - \Delta(V, S^+, S^-))/\delta)^2} - 1\right) \quad (7)$$

where $\delta$ determines the slope of the loss function; we set $\delta = \rho = 1.0$. The loss is differentiable because its derivatives are continuous for all degrees; this makes it especially attractive for learning with deep neural networks because gradient signals are more robust.

One way to compute the similarity $f(V, S)$ is to treat the embeddings $\Phi(V)$ and $\Psi(S)$ as $Kd$-dimensional vectors and compute:

$$f(V, S) = \frac{\Phi(V) \cdot \Psi(S)}{\|\Phi(V)\| \|\Psi(S)\|} \quad (8)$$

where $\cdot$ denotes the dot product between two vectors.

Note that this way of computing the similarity requires each of the $K$ video-sentence embedding pairs to be "aligned" in the shared embedding space; any misaligned dimension will negatively impact the similarity measure. This is problematic for our case because we assume the concepts in video-sentence pairs are implicitly associated with each other: It is natural to expect that only a few pairs of the embeddings to be aligned while others are severely misaligned. The form above, however, computes the dot product of the two embeddings, and thus has no ability to deal with potentially misaligned embedding pairs (the same reasoning applies even if $K = 1$). This motivates us to develop a MIL-based strategy, described next.

## 4.3 Multiple Instance Learning

In multiple instance learning (MIL), individual instances are assumed to be unlabeled; rather, the learner receives a set of labeled bags, each containing multiple instances with unknown (possibly different) labels [8]. For example, in binary classification all the instances in a negative bag are assumed to be negative, but only a few instances in a positive bag need to be positive. This provides flexibility in learning from weakly-supervised data and has shown to be successful in solving many real-world tasks [1].

To formulate our problem in the MIL framework, we start by assuming that each video and sentence contains multiple "concepts" that can be interpreted in different ways; consequently, a video-sentence pair has different combinations of visual-textual concepts (see Figure 1). Under this assumption, if a pair has *explicit* concept association, there should be no ambiguity in the interpretation of the pair, and thus every possible combination of visual-textual concepts should be valid. However, if the association is *implicit*, as is in our scenario, we assume ambiguity in the interpretation of the pair, and thus only a few visual-textual concept pairs need to be valid.

We define our "bag of instances" to contain all possible combinations of video-sentence embeddings from a pair $(V, S)$. Formally, we define a bag of instances $F(V, S)$ as

$$F(V, S) = \left\{ f_{i,j}(V, S) = \frac{\phi(V)_i \cdot \psi(S)_j}{\|\phi(V)_i\| \|\psi(S)_j\|} \right\}, \quad \forall i, j \in [1, \cdots, K] \quad (9)$$

Note that we have $K^2$ combinations of embeddings instead of $K$; this improves sample efficiency because any of $\Phi(V)$ can be matched up with any of $\Psi(S)$. We then modify Eqn. (6) to

$$\Delta_{mil}(V, S^+, S^-) = \max F(V, S^+) - \max F(V, S^-) \quad (10)$$

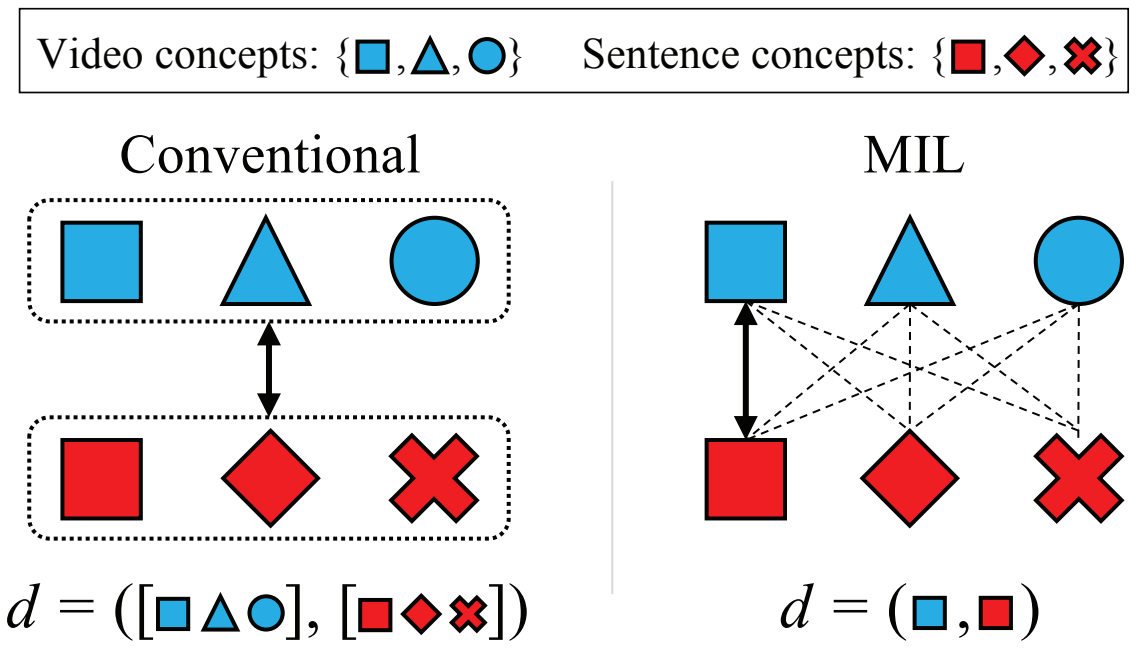

**Figure 6: In our scenario, each sample (video or sentence) is represented by $k$ embeddings, each capturing different concepts. Conventional approaches measure the visual-semantic distance by considering all $k$ embeddings (e.g. by concatenation; see Eqn.** (8)), **and would therefore suffer when not all concepts are related across modality. Our MIL-based approach, on the other hand, measures the distance from the most related concept combination (squares) by treating each embedding individually.**

and plug it into our loss function of Eqn. (7) to train our model.

Note that we have the max operator for both the positive and the negative bags. Similar to the case with MIL for binary classification [8], we require only a few instances in a positive bag to be valid. However, unlike [8], we also encourage only a few (rather than all) instances to be "invalid". This is intuitive and important: even if $V$ is paired with a randomly sampled $S^-$, some concept combinations of the two can still be valid. Our preliminary experiments showed that the performance is significantly worse if we do not use the max operator for the negative bags, which suggests that providing this "leeway" to negative bags is crucial in training our model.

## 4.4 Training and Inference

We train our model by minimizing the following objective function:

$$\min \sum_{n=1}^{N} \mathcal{L}_{mil-huber}(V_n, S_n^+, S_n^-) + \alpha \mathcal{R}(A_n) \tag{11}$$

where $\mathcal{L}_{mil-huber}(\cdot)$ is the pseudo-Huber loss (Eqn. (7)) with our MIL triplet ranking formulation (Eqn. (10)) and $\mathcal{R}(A_n) = \mathcal{R}(A_n^v) + \mathcal{R}(A_n^{s^+}) + \mathcal{R}(A_n^{s^-})$ is a regularization term that sets certain constraints to the attention maps for the $n$-th input triplet. The weight parameter $\alpha$ balances the influence between the two terms; we set $\alpha = 1e-4$.

Following [25], we design $\mathcal{R}(\cdot)$ to encourage $K$ attention maps for a given sequence to be diverse, i.e., each map attends to different parts of a sequence. Specifically, we define it as:

$$\mathcal{R}(A) = \|AA^\intercal - \beta I\|_F \tag{12}$$

where $\beta$ is a weight parameter, $I$ is a $K$-by-$K$ identity matrix, and $\|\cdot\|_F$ is the Frobenius norm. The first term promotes $K$ attention maps to be orthogonal to each other, encouraging diversity. The second term drives each diagonal term in $AA^\intercal$ to be close to $\beta \in [0, 1]$. In the extreme case of $\beta = 1$, each attention map is encouraged to be sparse and attend to a single component in a sequence (because its $l_2$ norm, the diagonal term, need to be 1). Lin *et al.* [25] suggests $\beta = 1$; however, we found that setting $\beta = 0.5$ leads to better performance.

For the inference (i.e., test time), we find the best matching video (or sentence) given a query sentence (or video) by computing cosine similarities between all $K^2$ combinations of embeddings (see Eqn (9)) and select the one with the maximum similarity. This has the time complexity of $O(K^2 N)$ with $N$ samples in the database.

# 5 EXPERIMENTS

## 5.1 Methodology

**Datasets**: We evaluate our approach on a sentence-to-video retrieval task using two datasets: the TGIF [23] and our MRW datasets. The former contains sentences describing visual content in videos, while the latter contains videos showing physical or emotional responses to sentences; therefore, the former contains *explicit* video-sentence concept association, while the latter contains *implicit* concept association. We split both datasets following the method of [23], which uses 80% / 10% / 10% of the data as the train / validation / test splits.

**Metrics**: We report our results using the median rank (MR) and recall@$k$ with $k = \{1, 5, 10\}$ (R@1, R@5, R@10). Both are widely used metrics for evaluating cross-modal retrieval systems [20, 37, 41]: the former measures the median position of the correct item in a ranked list of retrieved results, while the latter measures how often the correct item appears in the top $k$ of a ranked list. We also include normalized MR (nMR), which is the median rank divided by the total number of samples, so that it is independent of the dataset size. As both the datasets do not come with categorical labels, we do not use mean average precision (mAP) as our metric.

**Baselines**: We compare our method with DCCA [2], Corr-AE [12], and DeViSE [13]; these are well established methods used frequently in the cross-modal retrieval task. For fair comparisons, we use the same image and word embedding models (ResNet50 [17] and GloVe [29], respectively) as well as the same sequence encoder (Bidirectional GRU [6]) for all the models including the baselines.

## 5.2 Implementation Details

We set the maximum length of video and sentence to be 32 and zero-pad short sequences. For sequences longer than the maximum length, we sample multiple subsequences with random starting points and with random sampling rates. We apply dropout to the linear transformation for the input in the GRU, with a rate of 20% as suggested by [43]. We vary the number of GRU hidden units $d \in \{128, 256, 512\}$, the number of embeddings $K \in \{2, 4, 6, 8, 10, 12\}$, regularization weight $\lambda = 10^{-p}$ with $p \in \{1, 2, 3, 4\}$, and report the best results based on cross-validation; we choose the best model based on the nMR metric. We use the ADAM optimizer with a learning rate of 2e-4 and train our model for 500 epochs with a batch size of 100. Our model is implemented in TensorFlow.

## 5.3 Results and Discussion

Table 2 summarizes the results. From the results, we make the following observations: (1) Our approach outperforms all three baselines on both datasets, suggesting its superiority. (2) The overall performance on the TGIF dataset is better than on our dataset. This suggests the difficulty of our task with implicit concept association. (3) DeViSE and MiViSE perform better than DCCA and Corr-AE; the former are triplet ranking approaches, while the latter are correlation-based approaches. This suggests the superiority of the triplet ranking approach on sentence-to-video retrieval tasks. (3) DCCA performs poorly on both datasets. We believe this is due to the difficulties with estimating the covariance matrix required for DCCA.

| Method | TGIF Dataset [23] | | | | MRW Dataset (Ours) | | | |
|---|---|---|---|---|---|---|---|---|
| | MR (nMR) | R@1 | R@5 | R@10 | MR (nMR) | R@1 | R@5 | R@10 |
| DCCA [2] | 3722 (33.84) | 0.02 | 0.08 | 0.17 | 2231 (44.62) | 0.02 | 0.12 | 0.20 |
| Corr-AE [12] | 1667 (15.02) | 0.05 | 0.26 | 0.53 | 2181 (43.62) | 0.01 | 0.12 | 0.28 |
| DeViSE [13] | 861 (7.69) | 0.16 | 1.04 | 1.72 | 1806 (36.13) | 0.08 | 0.28 | 0.50 |
| **MiViSE** (Ours) | **426 (3.77)** | **0.58** | **2.00** | **3.74** | **1578 (31.57)** | **0.14** | **0.38** | **0.64** |

**Table 2: Sentence-to-video retrieval results on the TGIF and the MRW datasets. Metrics: median rank (MR), normalized MR (nMR), recall @ $k = \{1, 5, 10\}$ (R@1, R@5, R@10). For MR and nMR, the lower the better; for R@$k$, the higher the better.**

| MIL | SA | ME | MR (nMR) | R@1 | R@5 | R@10 |
|---|---|---|---|---|---|---|
| DeViSE [13] | | | 1806 (36.13) | 0.08 | 0.28 | 0.50 |
| ✗ | ✗ | ✗ | 1747 (34.94) | 0.02 | 0.16 | 0.46 |
| ✗ | ✓ | ✗ | 1693 (33.86) | 0.08 | 0.34 | **0.66** |
| ✗ | ✓ | ✓ | 1714 (34.29) | 0.10 | 0.34 | 0.52 |
| ✓ | ✓ | ✓ | **1578 (31.57)** | **0.14** | **0.38** | 0.64 |

**Table 3: Detailed evaluation results on the MRW dataset. Ablative factors are: multiple instance learning (MIL), self-attention (SA), multiple embeddings (ME). See the text for detailed explanation about different settings.**

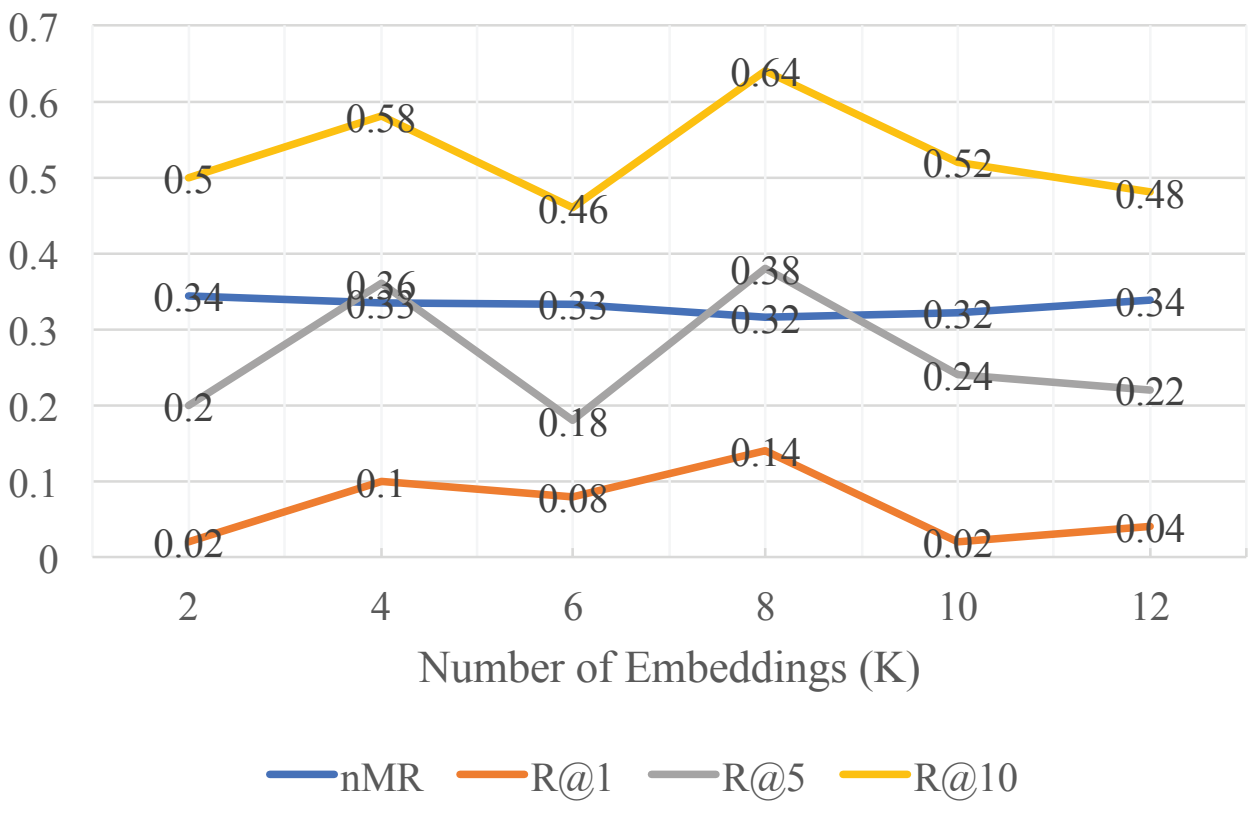

**Figure 7: Performance on the MRW dataset with different numbers of embeddings (K). The best results are obtained at $K = 8$.**

Both our approach and DeViSE are trained using a triplet ranking loss; but there are three major differences – (i) we replace the hinge loss (Eqn. (5)) to the pseudo-Huber loss (Eqn (7)), (ii) we compute multiple embeddings per modality using self-attention (Eqn. (4)), and (iii) we train our model in the framework of multiple instance learning (Eqn. (10)). To tease apart the influence of each of these factors, we conduct an ablation study adding one feature at a time; the results are shown in Table 3. We also include the DeViSE results from Table 2 for easy comparison.

From the results we make the following observations: (1) **Rows 1 vs. 2**: The only difference between DeViSE and ours without the three features is the loss function: the hinge loss vs. the pseudo-Huber loss. We can see that just by replacing the loss function we obtain a relative 3.29% performance improvement in terms of nMR. This shows the importance of having a fully differentiable loss function in training deep neural networks. (2) **Rows 2 vs. 3**: Adding self-attention provides a relative 3.09% performance improvement over the base model in nMR. Note that the two methods produce embeddings with the same feature dimensionality, $\mathbb{R}^d$; the former concatenates the two last hidden states of bidirectional GRUs, while the latter computes a linear combination of the hidden states using a single attention map (i.e., $K = 1$). Our result conforms to the literature showing the superiority of an attention-based representation [27, 40]. (3) **Rows 3 vs. 4**: Computing multiple embeddings and treating them as a single representation to compute the similarity (i.e., Eqn. (8)) hurts the performance. This is because each of the $K$ embeddings are required to be aligned in the embedding space, which might be a too strict assumption for the case with implicit concept association. (4) **Rows 4 vs. 5**: Training our model with MIL provides a relative 7.93% performance improvement in nMR. Note that the two methods produce embeddings with the same feature dimensionality, $\mathbb{R}^{Kd}$; the only difference is in how we compute the similarity between two embeddings, i.e., Eqn. (8) vs. Eqn. (9). This suggests the importance of the MIL in our framework.

Figure 7 shows the sensitivity of our MiViSE model to different numbers of embeddings; the results are based on our MRW dataset. We divide the nMR values by 100 for better visualization. We analyze the results in terms of the nMR metric because we choose our best model based on that metric. At $K = 2$, the model performs worse (nMR = 34.37) than our ablative settings (row 3 and 4 in Table 3). The performance improves as we increase $K$ and plateaus at $K = 8$ (nMR = 31.57). We can see that within a reasonable range of $2 < K < 12$, our model is insensitive to the number of embeddings: the relative difference between the best ($K = 8$, nMR = 31.57) and the worst ($K = 4$, nMR = 33.4) is 5.6%.

Finally, we visualize some examples of visual-textual attention maps in Figure 8. We see that attention maps for sentences change depending on the corresponding video, e.g., in the first row, the highest attention is given to the word "witty" in case of the predicted best matching video, while in case of the ground truth video it is given to the word "already". We also see evidence of multiple embeddings capturing different concepts. In the second row, the word "guy" is highlighted for the video with a male hockey player; when combined with "how he feels about me", this may capture such concepts as awkwardness. On the other hand, in case of the ground truth video, only the phrase "feels about me" is highlighted, which may capture different concepts, such as fondness.

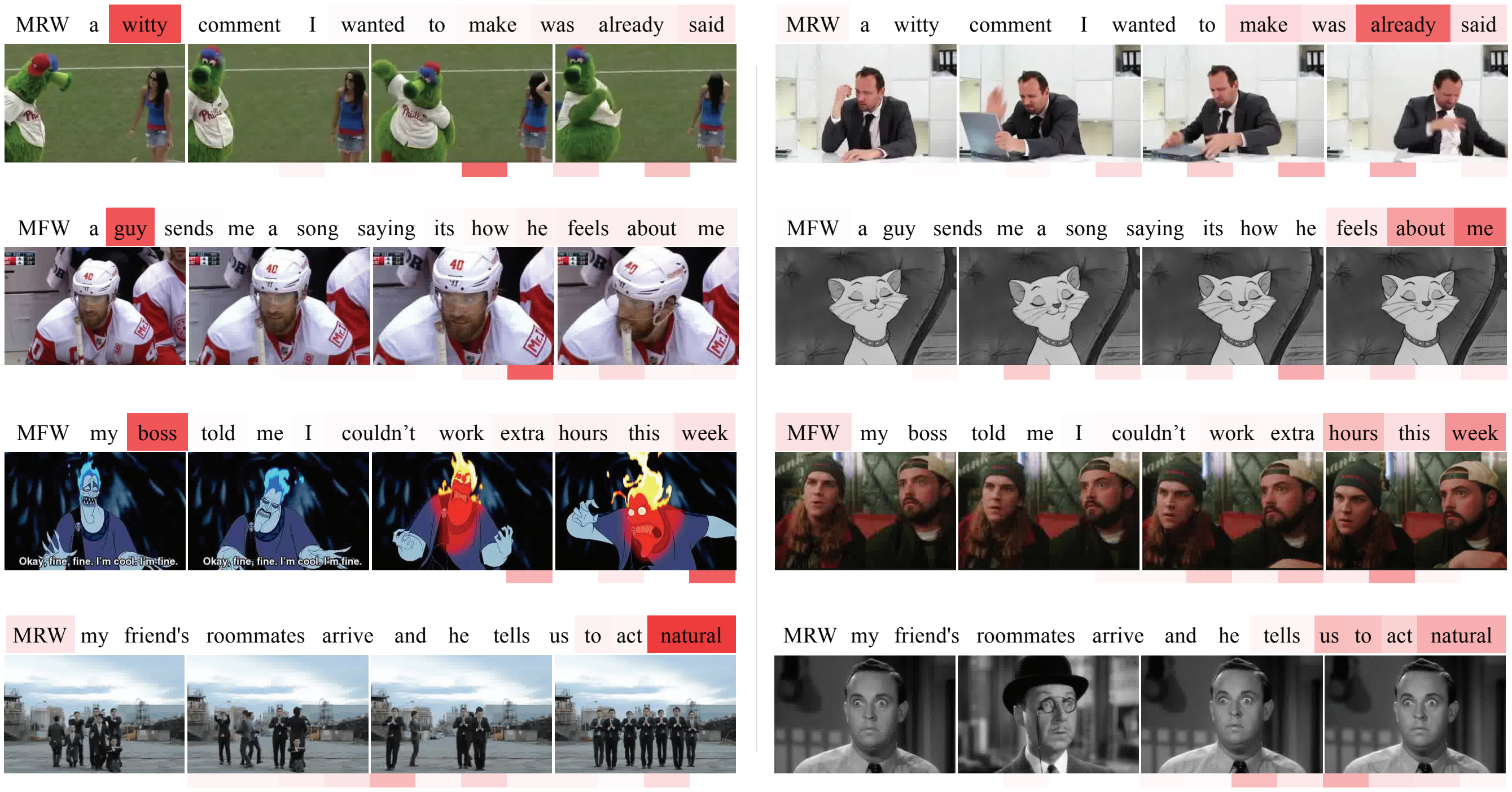

(a) Predicted best match (b) Ground truth

**Figure 8: Attention maps for (a) predicted best matching pairs and (b) the ground truth pairs; note the same sentence in each row. We show four frames uniformly sampled from each GIF; visual attention maps are shown below the frames. See the text for discussion.**

## 6 LIMITATIONS AND FUTURE WORK

**Using extra information**: We demonstrated our proposed MiViSE network on sentence-to-GIF retrieval, which is motivated by the real-world scenario of GIF search. For simplicity, we used generic video and sentence representation (CNN and GloVe, respectively, followed by a Bi-RNN); we did not leverage the full potential of our MRW dataset for this task. We believe that considering other sources of information available in our dataset, such as facial expression and captions in GIFs, will further improve the performance.

**Multiple embeddings**: Our MIL-based approach requires multiple embeddings from each modality, each capturing different concepts. We used self-attention to compute multiple embeddings, each attending to different parts of a sequence. While each embedding is encouraged to attend to different parts of a sequence, this does not guarantee that they capture different "concepts" (even though we see some evidence in Figure 8). We believe that explicitly modeling concepts via external resources could further improve the performance, e.g., using extra datasets focused on emotion and sentiment.

**Evaluation metric**: Our user study shows a high false positive rate of 50.7% when there is implicit concept association. This means there could be many "correct" videos for a sentence query. This calls for a different metric that measures the perceptual similarity between queries and retrieved results, rather than exact match. There has been some progress on perceptual metrics in the image synthesis literature (e.g., Inception Score [34]). We are not aware of a suitable perceptual metric for cross-modal retrieval, and this could be a promising direction for future research.

**Data size**: Our dataset is half the size of the TGIF dataset [23]. This is due to the real-world limitation: our data must be crawled, rather than generated by paid crowdworkers. We are continuously crawling the data, and plan to release updated versions on a regular basis.

## 7 CONCLUSION

In this paper, we addressed the problem of cross-modal multimedia retrieval with implicit association. For this purpose, we collected a dataset of reaction GIFs that we call MRW and their associated sentences from the Internet. We compared the characteristics of MRW dataset with the closest existing dataset developed for video captioning, i.e, TGIF. We proposed a new visual-semantic embedding network based on multiple instance learning and showed its effectiveness for cross-modal retrieval. Having a diverse set of attention maps in addition to using multiple instance learning, our method is better able to deal with the problem of alignment in the context of implicit association in reaction GIF retrieval. Our results demonstrated that the proposed method outperforms correlation-based methods, i.e., Corr-AE [12] and DCCA [2], as well as existing methods with triplet ranking, i.e, DeViSE [13]. We carefully compared different variations of our model and showed factors that contribute to the improved performance. Finally, we discussed limitations and directions for future work, some of which we have started to work on.